\def\year{2022}\relax
\documentclass[letterpaper]{article} 
\usepackage{aaai22}  
\usepackage{times}  
\usepackage{helvet}  
\usepackage{courier}  
\usepackage[hyphens]{url}  
\usepackage{graphicx} 
\usepackage{natbib}  
\usepackage{caption} 
\DeclareCaptionStyle{ruled}{labelfont=normalfont,labelsep=colon,strut=off} 
\usepackage{algorithm}
\usepackage{algorithmic}
\usepackage{color}
\usepackage{newfloat}
\usepackage{listings}
\floatstyle{ruled}
\newfloat{listing}{tb}{lst}{}
\floatname{listing}{Listing}
\usepackage{amsmath}
\usepackage{amssymb}
\usepackage{multirow}

\title{DarkVisionNet: Low-Light Imaging via RGB-NIR Fusion \\with Deep Inconsistency Prior}
\author {
    Shuangping Jin,\textsuperscript{\rm 1, \rm 2}\thanks{Authors contributed equally.}
    Bingbing Yu, \textsuperscript{\rm 1, \rm 3}\footnotemark[1]
    Minhao Jing \textsuperscript{\rm 1}
    Yi Zhou \textsuperscript{\rm 1}
    Jiajun Liang \textsuperscript{\rm 1}
    Renhe Ji \textsuperscript{\rm 1}\thanks{Corresponding author.}
}
\affiliations {
    \textsuperscript{\rm 1} Megvii Technology \\ 
    \textsuperscript{\rm 2} Southeast University \\ 
    \textsuperscript{\rm 3} Dalian University Of Technology \\ 
    220191583@seu.edu.cn, 21901037@mail.dlut.edu.cn, jingminhao@megvii.com, zhouyi@megvii.com, liangjiajun@megvii.com, jirenhe@megvii.com
}

\usepackage{bibentry}

\begin{document}

\maketitle

\begin{abstract}
RGB-NIR fusion is a promising method for low-light imaging. 
However, high-intensity noise in low-light images amplifies the effect of structure inconsistency between RGB-NIR images, which fails existing algorithms. 
To handle this, we propose a new RGB-NIR fusion algorithm called Dark Vision Net (DVN) with two technical novelties: Deep Structure and Deep Inconsistency Prior (DIP). 
The Deep Structure extracts clear structure details in deep multiscale feature space rather than raw input space, which is more robust to noisy inputs. 
Based on the deep structures from both RGB and NIR domains, we introduce the DIP to leverage the structure inconsistency to guide the fusion of RGB-NIR. 
Benefiting from this, the proposed DVN obtains high-quality low-light images without the visual artifacts. 
We also propose a new dataset called Dark Vision Dataset (DVD), consisting of aligned RGB-NIR image pairs, as the first public RGB-NIR fusion benchmark. Quantitative and qualitative results on the proposed benchmark show that DVN significantly outperforms other comparison algorithms in PSNR and SSIM, especially in extremely low light conditions. Code and Data can be found in https://github.com/megvii-research/DVN.
\end{abstract}

\section{Introduction}

\begin{figure}[!t]
	\centering
	\includegraphics[width=0.8\linewidth]{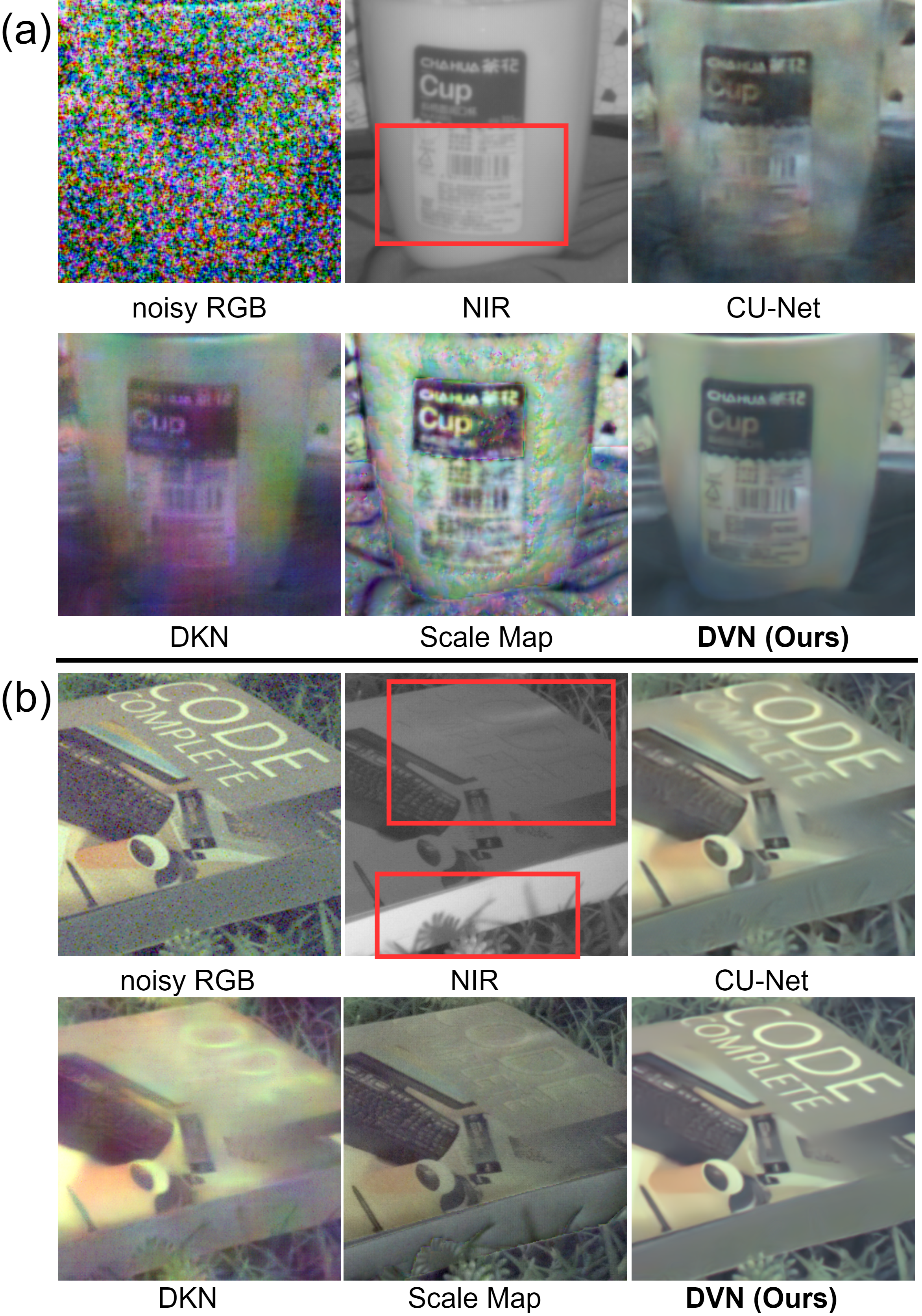}
	\caption{(a) and (b) are fusion examples from DVD, compared to CU-Net \cite{deng2020deep}, DKN \cite{kim2021deformable} and Scale Map \cite{yan2013cross}, our method, Dark Vision Net (DVN), effectively handle the structure inconsistency between RGB-NIR images. Regions with inconsistent structures are framed in red.}
	\label{fig:intro}
	\vspace{-1em}
\end{figure}

High-quality low-light imaging is a challenging but significant task. On the one hand, it is the cornerstone of many important applications such as 24-hour surveillance, smartphone photography, etc. On the other hand, though, massive noise of images in extremely dark environments hinders algorithms from the satisfactory restoration of low-light images. RGB-NIR fusion techniques provide a new perspective for the challenge: it enhances the low-light noisy color (RGB) image through rich, detailed information in the corresponding near-infrared (NIR) image (The high quality of NIR images in dark environments comes from invisible near-infrared flash), which greatly improves the signal-to-noise ratio (SNR) of the restored RGB image. Under the constraint of cost, size and other factors, RGB-NIR fusion becomes the most promising technique to restore the vanished textual and structural details from noisy RGB images taken in an extremely low-light environments, as shown in Figure \ref{fig:intro}(a).

\begin{figure*}[!t]
	\centering
	\includegraphics[width=0.8\linewidth]{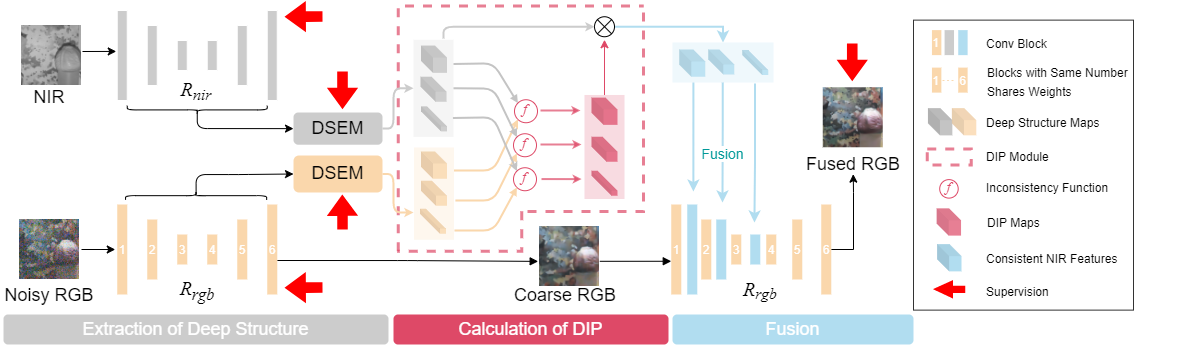}
	\caption{Overview of the proposed DVN. In the first stage, the network predicts deep structure maps utilising the multi-scale features maps from restoration network $R$ by the proposed Deep Structure Extraction Module (DSEM) for noisy RGB and NIR respectively. In the second stage, taking advantage of the predicted deep structures, the DIP can be calculated by inconsistency function $\mathcal{F}$. In the third stage, the DIP-weighted NIR structures are fused with the RGB features to obtain the final fusion result without obvious structure inconsistency. }
	\label{fig:archi}
	\vspace{-1em}
\end{figure*}

However, the existing RGB-NIR fusion algorithms suffers from the problem of structure inconsistency between RGB and NIR images, resulting in unnatural appearance and loss of key information, which limits the application of RGB-NIR fusion algorithm in low-light imaging. 
Figure \ref{fig:intro} illustrates two typical examples of structure inconsistency between RGB and NIR images: Figure \ref{fig:intro}(b) shows the absence of NIR shadows in the RGB image (grass shadows only appear on the book edge in the NIR image), and the nonexistence of RGB color structure in the NIR image (text ‘complete’ almost disappears on the book cover in the NIR image). Fusion algorithms need to tackle these structure inconsistency to avoid visual artifacts in output images. There are two categories of RGB-NIR fusion methods currently, \textit{i.e.} traditional methods and neural-network-based methods, and modeling the structure of the paired RGB-NIR images plays an important role in both of them. Traditional methods, such as Scale Map \cite{yan2013cross}, tackle the structure inconsistency problem by manually designed functions. Some neural-network-based methods \cite{kim2021deformable,li2019joint}, on the other hand, utilize deep learning techniques to automatically learn the structure inconsistency by a large amount of data. Both of them perform well under certain circumstances.

However, when confronted with extreme low-light environments, existing methods fail to maintain satisfactory performance, since the structure inconsistency is dramatically exacerbated by massive noise in the RGB image. As shown in Figure \ref{fig:intro}(b), the dense noise in the RGB image makes it difficult for Scale Map to extract structural information, causing the failure of distinguishing which structures in the NIR image should be eliminated, and result in the unnatural ghost images on the book edge. Deformable Kernel Networks (DKN) \cite{kim2021deformable} falsely weakens gradients of input RGB image that do not exist in the corresponding NIR image, which leads to the blurriness of letters on the book cover. Even though these structural inconsistency of corresponding RGB and NIR images can be captured by human eyes effortlessly, they still confuse most of the existing fusion algorithms.

In this paper, we focus on improving the RGB-NIR fusion algorithm for extremely low SNR images by tackling the structure inconsistency problem. Based on the above analysis, we argue that the structure inconsistency under extremely low light can be handled well by introducing prior knowledge into deep features. To achieve this, we propose a deep RGB-NIR fusion algorithm called Dark Vision Net (DVN), which explicitly leverages the prior knowledge of structure inconsistency to guide the fusion of RGB-NIR deep features, as shown in Figure \ref{fig:archi}. With DVN, two technical novelties are introduced: (1) We find a new way, referred to as deep structures, to represent clear structure information encoded in deep features extracted by the proposed Deep Structure Extraction Module (DSEM). Even facing images with low SNR, the deep structures can still be effectively extracted and represent reliable structural information, which is critical to the introduction of prior knowledge. (2) We propose Deep Inconsistency Prior (DIP), which indicates the differences between RGB-NIR deep structures. Integrated into the fusion of RGB-NIR deep features, the DIP empowers the network to handle the structure inconsistency. Benefiting from this, the proposed DVN can obtain high-quality low-light images.

In addition, to the best of our knowledge, there is no available benchmark dedicated for the RGB-NIR fusion task so far. The lack of benchmarks to evaluate and train fusion algorithms greatly limits the development of this field. To fill this gap, we propose a dataset named Dark Vision Dataset (DVD) as the first RGB-NIR fusion benchmark. Based on this dataset, we give qualitative and quantitative evaluations to prove the effectiveness of our method.
In summary, the main contributions of this paper are as follows:

\begin{itemize}
	\item We propose a novel RGB-NIR fusion algorithm called Dark Vision Net (DVN) With Deep Inconsistency Prior (DIP). The DIP explicitly integrates the prior of structure inconsistency into the deep features, avoiding over-relying on NIR features in the feature fusion. Benefits from this, DVN can obtain high-quality low-light images without visual artifacts.
	\item We propose a new dataset Dark Vision Dataset (DVD) as the first public dataset for training and evaluating RGB-NIR fusion algorithms.
	\item The quantitative and qualitative results indicate that DVN is significantly better than other compared methods.
\end{itemize}

\vspace{-1em}
\section{Related Work}
\subsubsection{Image Denoising. }
In recent years, denoising algorithms based on deep neural networks have continually emerged and overcome the drawbacks of analytical methods \cite{lucas2018using}. The image noise model is gradually improved simultaneously \cite{wei2020physics, wang2020practical}. 
\cite{mao2016image} applied an encoder-decoder network to suppress the noise and recover high-quality images. 
\cite{zhang2018ffdnet} presented a denoising network to process blind noise denoising.
\cite{guo2019toward, cao2021pseudo} attempted to remove the noise from real noisy images.
There are also deep denoising algorithms trained without clean data supervision \cite{lehtinen2018noise2noise, krull2019noise2void, huang2021neighbor2neighbor}. 
However, in extremely dark environments, fine texture details damaged by the high-intensity noise are very difficult to restore. In that case, denoising algorithms tend to generate over-smoothed outputs. By the way, low-light image enhancement algorithms \cite{chen2018learning, lamba2021restoring, gu2019self} try to directly restore high-quality images in terms of brightness, color, etc. However, these algorithms cannot deal with such high-intensity noise as well. 

\subsubsection{RGB-NIR Fusion. }
To obtain high-quality low-light images, researchers \cite{krishnan2009dark} try to fuse NIR images with RGB images. 
Recently, \cite{yan2013cross} pointed out the gradient inconsistency between RGB-NIR image pairs, and proposed Scale Map to try to solve it. 
Among the methods based on deep neural network, Joint Image Filtering with Deep Convolutional Networks (DJFR) \cite{li2019joint} constructs a unified two-stream network model for image fusion, CU-Net \cite{deng2020deep} combines sparse encoding with Convolutional Neural Networks (CNNs), DKN \cite{kim2021deformable} explicitly learns sparse and spatially-variant kernels for image filtering. \cite{lv2020integrated} innovatively constructs a network that directly decouples RGB and NIR signals for 24-hour imaging. 
In general, current RGB-NIR fusion algorithm has two main problems: insufficient ability to deal with RGB-NIR texture inconsistency, leading to heavy artifacts on the final fusion images, inadequate noise suppression capability especially when dealing with high-intensity noise in extremely low-light environments. 

\subsubsection{Datasets. }
There is only a small amount of data that can be used for RGB-NIR fusion studies because of the difficulty to obtain aligned RGB-NIR image pairs. Some studies \cite{foster2006frequency} focus on obtaining hyperspectral datasets, and strictly aligned RGB-NIR image pairs can be obtained by integrating hyperspectral images on the corresponding band. 
\cite{krishnan2009dark} present a prototype camera to collect RGB-NIR image pairs. However, these datasets are too small to be used to comprehensively measure the performance of fusion algorithms. 
More importantly, due to the lack of data on actual scenarios, they cannot encourage follow-up researchers to focus on the valuable problems that RGB-NIR will encounter in applications. 

\section{Approach}

\subsection{Prior Knowledge of Structure Inconsistency}
As previously described, the network needs to be aware of the inconsistent regions on the two inputs. We design an intuitive function to measure the inconsistency from image features. Firstly binary edge maps are extracted from each feature channel. Then the inconsistency is defined as
\begin{align}
\mathcal{F} (edge^{C_{:}}, edge^{N}) &= \lambda (1 - edge^{C_{:}})(1 - edge^{N}) \notag \\ & \quad + edge^{C_{:}} \cdot edge^{N}
\end{align}
where $C_{:} \in \mathbb{R}^{H \times W}$ and $N \in \mathbb{R}^{H \times W}$ denote R/G/B channel of the clean RGB image and NIR image, $edge^{C_{:}}$ and $edge^{N}$ respectively represent the binarized edge maps of $C_{:}$ and $N$, which is obtained by binarizing its mean value as a threshold after Sobel filtering. 

\begin{figure}[!t]
	\centering
	\includegraphics[width=\linewidth]{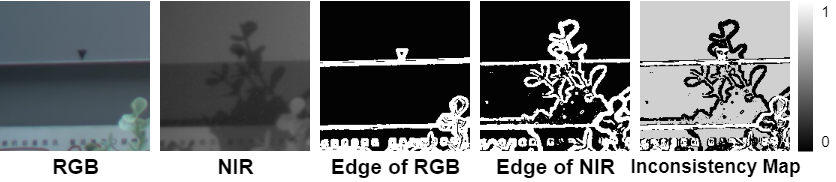}
	\caption{Through applying $\mathcal{F}$ on edge maps of clean RGB and NIR images, the calculated inconsistency map clearly shows the structure inconsistency between RGB and NIR.}
	\label{fig:motivation}
	\vspace{-1em}
\end{figure}

As shown in Figure \ref{fig:motivation}, $\mathcal{F}(\cdot,\cdot)$ equals to $0$ in the regions where $edge^{C_:}$ and $edge^N$ shows severe inconsistency. 
On the contrary, $\mathcal{F}(\cdot,\cdot)$ equals to $1$ in the regions where the structures of RGB and NIR are consistent. 
And in other regions, $\mathcal{F}(\cdot,\cdot)$ is set to a hyperparameter $\lambda (0 < \lambda < 1)$, indicating that there is no significant inconsistency. 
Utilising the output inconsistency map of $\mathcal{F}$, the inconsistent NIR structures can be easily suppressed by a direct multiplication.


\subsection{Extraction of Deep Structures}
\begin{figure}[!t]
	\centering
	\includegraphics[width=\linewidth]{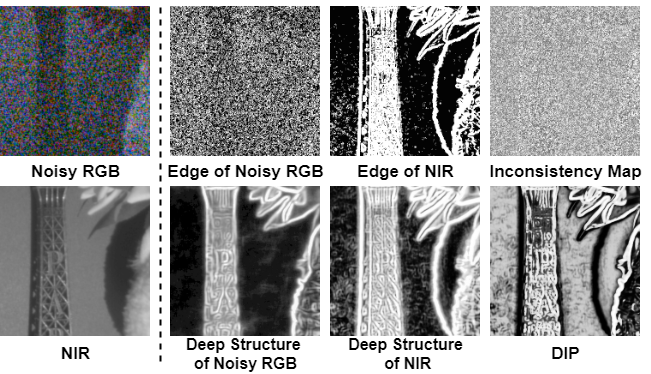}
	\caption{Applying $\mathcal{F}$ on the edge maps of noisy RGB and NIR images can only get a meaningless inconsistency map due to the heavy noise in the input RGB (as the first row shows). However, the deep structure map predicted by DSEM is very clear and the calculated DIP effectively describes the inconsistent structures (as the second-row shows). See more examples in supplementary material.}
	\label{fig:introdip}
	\vspace{-1em}
\end{figure}

Even though function $\mathcal{F}$ subtly describes the inconsistency between RGB and NIR images, it cannot be applied directly in extremely low light cases. 
As shown in Figure \ref{fig:introdip}, the calculated inconsistency map contains nothing but non-informative noise when facing extremely noisy RGB image. 
To avoid the influence of noise in the structure inconsistency extraction, we propose the Deep Structure Extraction Module (DSEM) and Deep Inconsistency Prior (DIP), where we compute the structure inconsistency in feature space. 
Considering the processing flow of RGB and NIR are basically the same, we give a unified description here to keep the symbols concise. 

\begin{figure*}[!ht]
	\centering
	\includegraphics[width=\linewidth]{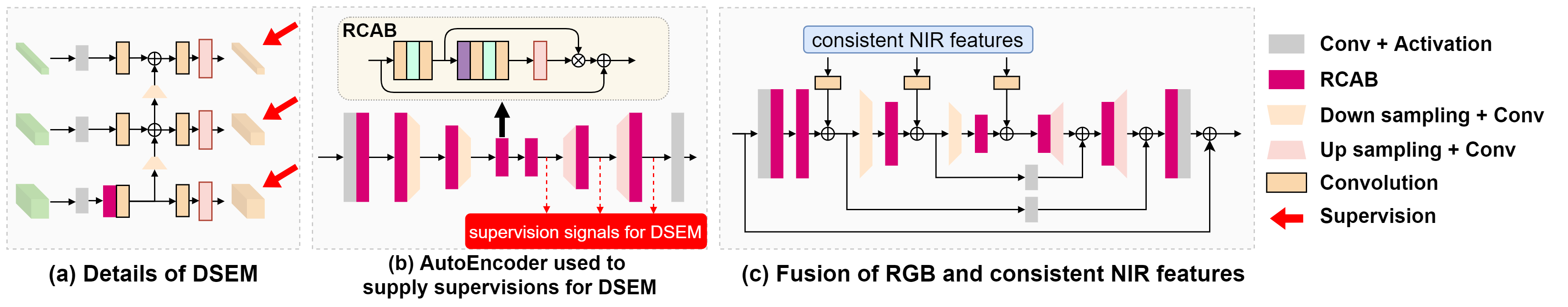}
	\caption{(a) Illustration of the Deep Structure Extraction Module (DSEM) details.(b) The architecture of the AutoEncoder network which is employed to provide supervision signals for DSEM. (c) The detailed fusion process of RGB and DIP-weighted NIR features, \textit{i.e.} the third stage of DVN. Residual channel attention blocks (RCABs) \cite{zhang2018image} are used to extract features. }
	\label{fig:approachdetail}
	\vspace{-1em}
\end{figure*}

The detailed architecture of DSEM is shown in Figure \ref{fig:approachdetail}(a). 
DSEM takes multi-scale features $feat_i$ ($i$ represents scale) from restoration network $R$ and outputs multi-scale deep structure maps $struct_i$. 
In order for DSEM to predict high-quality deep structure maps, we introduce a clear supervision signal $struct_{i}^{gt}$ (addressed later) for DSEM and the training loss is calculated as:
\begin{align}
    \mathcal{L}_{stru} = \sum_{i = 1}^{3} \sum_{c = 1}^{Ch_i} Dist(struct_{i, c}, struct_{i, c}^{gt}), 
\end{align}
where $Ch_i$ is the channel number of the deep structures in the $i$th scale, $Dist$ is Dice Loss \cite{deng2018learning}, $struct_{i, c}$ is the $c$th channel of the predicted deep structures in the $i$th scale and $struct_{i, c}^{gt}$ is the corresponding ground-truth. The Dice loss is given by
$ Dist(P, G) = ( \sum_{j}^{N} p_{j}^{2}+\sum_{j}^{N} g_{j}^{2} ) / (2 \sum_{j}^{N} p_{j} g_{j}), $ where $p_{j}$, $g_{j}$ is the value of the $j$th pixel on the predicted structure map $P$ and ground-truth $G$. 

\subsubsection{Supervision of DSEM}
Considering that it is almost impossible for DSEM to naturally output feature maps that only contain structural information, we have to introduce a clear supervision for the output of DSEM to predict high-quality deep structure maps. 
The supervision signal is set up following the idea of Deep Image Prior \cite{ulyanov2018deep} and $struct_{i, c}^{gt}$ is acquired from a pretrained AutoEncoder \cite{hinton2006reducing} network $\mathrm{AE}$ \footnote{See the training details in supplementary material.}. 
The base architecture of $\mathrm{AE}$ is exactly the same as $R$ with skip connections removed, as Figure \ref{fig:approachdetail}(b) shows. 
Multi-scale decoder features $dec_{i, c}$ are extracted from the pretrained AutoEncoder network $\mathrm{AE}$ and the supervision signal is calculated by:
\begin{equation} struct_{i, c, j}^{gt} =
\begin{cases}
0 & \text{if } (\nabla dec_{i, c, j} - m_{\nabla dec_{i, c}}) <= 0,\\
1 & \text{if } (\nabla dec_{i, c, j} - m_{\nabla dec_{i, c}}) > 0. 
\end{cases} 
\end{equation}
where $struct_{i, c, j}^{gt}$ is the $j$th pixel of $struct_{i, c}^{gt}$, $\nabla$ represents the Sobel operator, $\nabla dec_{i, c, j}$ is the $j$th pixel in $\nabla dec_{i, c}$ and $m_{\nabla dec_{i, c}}$ is the global average pooling result of $\nabla dec_{i, c}$. The supervision signal obtained by this design effectively trains the DSEM and clear deep structure maps are predicted as shown in Figure \ref{fig:introdip}.

\subsection{Calculation of DIP and Image Fusion}
The extracted deep structures contain rich structure information and are robust to noise. With $struct_{i}$ of the noisy RGB and NIR images, we can introduce inconsistency function $\mathcal{F}$ to obtain high-quality knowledge of structure inconsistency:
\begin{align}
    M^{DIP}_{i, c} &= \mathcal{F}(struct_{i, c}^C, struct_{i, c}^N)
\end{align}
where $C \in \mathbb{R}^{H \times W \times 3}$ and $N \in \mathbb{R}^{H \times W}$ denote the noisy RGB image and NIR image, $struct_{i, c}$ is the $c$th channel of the features from the $i$th scale and $M^{DIP}_{i, c}$ is the corresponding inconsistency measurement. 
Since $M^{DIP}_{i, c}$ represents the structure inconsistency instead of intensity inconsistency, we apply $M^{DIP}_{i, c}$ directly to $struct_{i, c}^N$ instead of $feat^{N}_{i, c}$ in the form of:
\begin{equation}
    \hat{struct_{i, c}^N} = M^{DIP}_{i, c} \cdot struct_{i, c}^N. 
\end{equation}
Under the guidance of the DIP, $\hat{struct_{i, c}^N}$ discards the structures that are inconsistent with RGB, thus empowering the deep features with prior knowledge to tackle structure inconsistency. 
As we will show in the experiments later, inconsistent structures in the NIR structure maps can be significantly suppressed after multiplying with $M^{DIP}_{i, c}$. 

\begin{figure*}[!t]
	\centering
	\includegraphics[width=.8\linewidth]{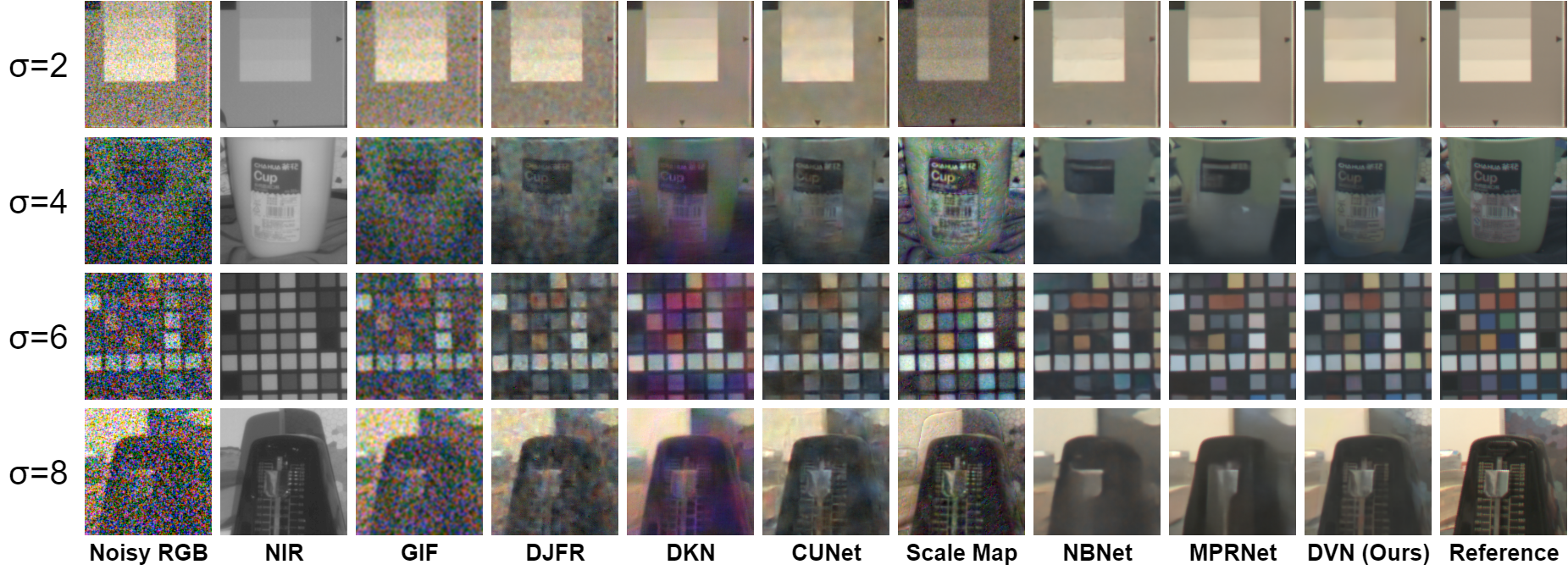}
	\caption{Fusion examples from DVD. The proposed DVN shows great superiority than other algorithms. Images are brightened for visual convenience. See supplementary material for more examples. }
	\label{fig:performCompare}
	\vspace{-1em}
\end{figure*}

To further fuse the rich details contained in $\hat{struct_{i, c}^N}$ into RGB features, we designed a multi-scale fusion module as shown in Figure \ref{fig:approachdetail}(c). As pointed out in \cite{jung2020fusionnet}, denoising first and fusion later can improve the fusion quality. So we follow \cite{jung2020fusionnet} to reuse the denoised output of the restoration network $R$ set up for noisy RGB as the input of the multi-scale fusion module.
\vspace{-1em}
\subsection{Loss Function}
The total loss function we used is formulated as:
\begin{equation}
\mathcal{L} = \mathcal{L}_{rec}^{C} + \mathcal{L}_{rec}^{\hat{C}} + \mathcal{L}_{rec}^{N} + \lambda_1 \cdot \mathcal{L}_{stru}^{C} + \lambda_2 \cdot \mathcal{L}_{stru}^{N}
\end{equation}
where $\mathcal{L}_{stru}^{C}$ and $\mathcal{L}_{stru}^{N}$ are the loss function for RGB/NIR deep structures prediction, which is described above. $\lambda_1$ and $\lambda_2$ are the corresponding coefficients and set to $1/1000$ and $1/3000$. 
$\mathcal{L}_{rec}^{C}$, $\mathcal{L}_{rec}^{\hat{C}}$ and $\mathcal{L}_{rec}^{N}$ represent the reconstruction loss of fused-RGB/coarse-RGB/NIR image respectively. All of them are Charbonnier loss \cite{charbonnier1994two} in the form of:
\begin{align}
    \mathcal{L}_{rec} &= \sqrt{\left\| \mathbf{X}-\mathbf{X}_{gt}\right \|^{2}+\varepsilon^{2}}
\end{align}
where $\mathbf{X}$ and $\mathbf{X}_{gt}$ represent the network output and the corresponding supervision. The constant $\varepsilon$ is set to $10^{-3}$ empirically. 


\section{Experiment}

\subsection{Datasets}

\subsubsection{Data Collection. }
In order to obtain the aligned RGB-NIR image pairs in the easiest and direct way, we collect all RGB-NIR image pairs by switching an optical filter placed directly in front of the camera without an IR-Cut, and we divide them into two types of image pairs for different usages.
We collect \textbf{reference image pairs} in normal-light environments. In order to obtain high-quality references, multiple still captures are averaged to remove noise and a matching algorithm \cite{superpoint} with manual double-check is applied to ensure the alignment of image pairs. In the following experiments, we add synthetic noise to these references to quantitatively evaluate the performance of fusion algorithms. To facilitate training, the collected images are cropped into 256*256 image patches.
We collect \textbf{real noisy image pairs} of 1920*1080 pixels in low-light environments. The post-processing steps are the same as those used in collecting references image pairs, except that multi-frame average is not performed to noisy RGB images. In the following experiments, we use these noisy image pairs to qualitatively evaluate the performances of fusion algorithms in handling real low-light images.

\subsubsection{Dataset for Experiment. }
To make the synthetic data closer to the real images, we follow \cite{wang2020practical} to add synthetic noise to reference image pairs for training and testing. Considering that all the comparison methods use sRGB (standard Red Green Blue) images as input, we convert the collected raw data into sRGB through a simple isp-pipeline (Gray World for white balance, Gamma correction, Demosaicing) \cite{karaimer2016software}, to make a fair comparison. In the following experiments, we use 5k reference image pairs (256*256) as the training set. Another 1k reference image pairs (256*256) along with 10 additional real noisy image pairs (1920*1080) are used for testing. 

\begin{table*}[!ht]
\centering
\vspace{-0.5em}
\begin{tabular}{llcccccccc}
\hline
\multicolumn{2}{l}{}      & GIF & DJFR & DKN & Scale Map & CUNet & NBNet & MPRNet & DVN (Ours) \\ \hline
\multirow{2}{*}{$\sigma = 2$} & PSNR & 22.32    & 26.28     & 27.22    & 21.98      & 28.62          & 31.38      & \textbf{31.79}       & \textit{31.50}           \\ \cline{2-10} 
                  & SSIM & 0.6410    & 0.8263     & 0.8902    & 0.6616      & 0.9138     & 0.9477      & \textit{0.9504}       & \textbf{0.9551}        \\ \hline
\multirow{2}{*}{$\sigma = 4$} & PSNR & 19.15    & 23.91     & 24.34    & 21.02      & 26.81          & 29.14      & \textit{29.37}       & \textbf{29.62}           \\ \cline{2-10} 
                  & SSIM & 0.5033    & 0.7464     & 0.8427    & 0.6225      & 0.8832     & 0.9259      & \textit{0.9276}       & \textbf{0.9400}        \\ \hline
\multirow{2}{*}{$\sigma = 6$} & PSNR & 17.30    & 22.40     & 22.78    & 20.02      & 25.43          & 27.27      & \textit{27.68}       & \textbf{28.26}           \\ \cline{2-10} 
                  & SSIM & 0.4240    & 0.6802     & 0.8067    & 0.5959      & 0.8510     & 0.9060      & \textit{0.9083}       & \textbf{0.9273}        \\ \hline
\multirow{2}{*}{$\sigma = 8$} & PSNR & 15.98    & 20.72     & 22.50    & 19.07      & 23.75          & 24.81      & \textit{26.20}       & \textbf{26.98}           \\ \cline{2-10} 
                  & SSIM & 0.3701    & 0.6177     & 0.7799    & 0.5742      & 0.8154     & 0.8822      & \textit{0.8908}       & \textbf{0.9155}        \\ \hline
\end{tabular}
\caption{The PSNR (dB) and SSIM results of different algorithms on DVD dataset. The best and second best results are highlighted in bold and Italic respectively. }
\label{table:performancecomparison}
\end{table*}

\begin{table*}[!ht]
\centering
\begin{tabular}{cc|cc}
\hline
\multicolumn{2}{c|}{\begin{tabular}[c]{@{}c@{}}PSNR comparison on public \\ IVRG dataset ($\sigma$=50, input PSNR = 13.44)\end{tabular}} & \multicolumn{2}{c}{\begin{tabular}[c]{@{}c@{}}PSNR comparison with \\ other methods on DVD ($\sigma = 4$)\end{tabular}} \\ \hline
\multicolumn{1}{c|}{DJFR} & 23.35 & \multicolumn{1}{c|}{SID (Chen et al. 2018)} & 25.26  \\
\multicolumn{1}{c|}{CUNet} & 24.96 & \multicolumn{1}{c|}{SGN (Gu et al. 2019)} & 28.40  \\
\multicolumn{1}{c|}{Scale Map} & 25.59 & \multicolumn{1}{c|}{SSN (Dong et al. 2018)} & 13.72  \\
\multicolumn{1}{c|}{DVN (Ours)} & 30.43 & \multicolumn{1}{c|}{DVN (Ours)} & 29.62  \\ \hline
\end{tabular}
\caption{Performance comparison (PSNR). The conclusions are the same if SSIM is applied as the metric.}
\vspace{-0.5em}
\label{table:lowlight}
\end{table*}

\subsection{Implementation Details}
All experiments are conducted on a device equipped with two 2080-Ti GPUs. 
We train the proposed DVN from scratch in an end-to-end fashion. 
Batchsize is set to 16. 
Training images are randomly cropped in the size of 128*128, and the value range is [0, 1]. 
We augment the training data following MPRNet \cite{zamir2021multi}, including random flipping and rotating. 
Adam optimizer with momentum terms (0.9, 0.999) is applied for optimization. 
The whole network is trained for 80 epochs, and the learning rate is gradually reduced from 2e-4 to 1e-6. 
$\lambda$ in function $\mathcal{F}$ is set to 0.5 for all configurations. 
The AutoEncoder network used to provide supervision signals for DSEM is pretrained in the same way, except that it only trained for 5 epoches and the input is clean RGB and NIR images separately. See supplementary material for more training details. 

The synthesis of low-light data for training includes two steps. 
The first step is to reduce the average value of raw images taken under normal light to 5 (10-bit raw data). The second step is to add noise to the pseudo-dark raw images, including Gaussian noise with variance equals to $\sigma$, and Poisson noise with a level proportional to $\sigma$.

\subsection{Performance Comparison}
\subsubsection{Results on DVD Benchmark. }
We evaluate and compare DVN with representative methods in related fields, including single-frame noise reduction algorithms NBNet \cite{cheng2021nbnet} and MPRNet \cite{zamir2021multi}, joint image filtering algorithms GIF \cite{he2012guided}, DJFR \cite{li2019joint}, DKN \cite{kim2021deformable} and CUNet \cite{deng2020deep}, as well as Scale Map \cite{yan2013cross} which specially designed for RGB-NIR fusion. 
All methods are trained or finetuned on DVD from scratch. 
We use PSNR and SSIM \cite{wang2004image} for quantitative measurement. 
Qualitative comparison is shown in Figure \ref{fig:performCompare}, and quantitative comparison under different noise intensity settings ($\sigma = 2, 4, 6, 8$, the larger the $\sigma$, the heavier the noise) on DVD benchmark is shown in Table \ref{table:performancecomparison}. 

The qualitative comparison in Figure \ref{fig:performCompare} clearly illustrates the superiority of the proposed DVN on noise removal, detail restoration and visual artifacts suppression.
In contrast, image denoising algorithms (\textit{i.e.} NBNet and MPRNet) cannot restore texture details when the noise intensity becomes high, and the output turns into pieces of smear even though the noise is effectively suppressed. GIF and DJFR output images with heavy noise as the $3$rd and $4$th column in Figure \ref{fig:performCompare} shows, which greatly affects the fusion quality. The fusion effect of DKN and CUNet ($5$th and $6$th column in Figure \ref{fig:performCompare}) under mild noise (\textit{e.g.} $\sigma=2$) is acceptable. But under heavy noise, obvious color deviation appears in the DKN output, and neither of them can deal with structure inconsistency (see the 4th row in Figure \ref{fig:performCompare}), resulting in severe artifacts in the fusion images. Scale Map outputs images with rich details. However, it cannot reduce the noise in the areas where texture is lacking in the NIR image. In addition, it is hard to achieve a balance between noise suppression and texture migration when applying Scale Map. 




\begin{figure*}[!t]
	\centering
	\includegraphics[width=0.8\linewidth]{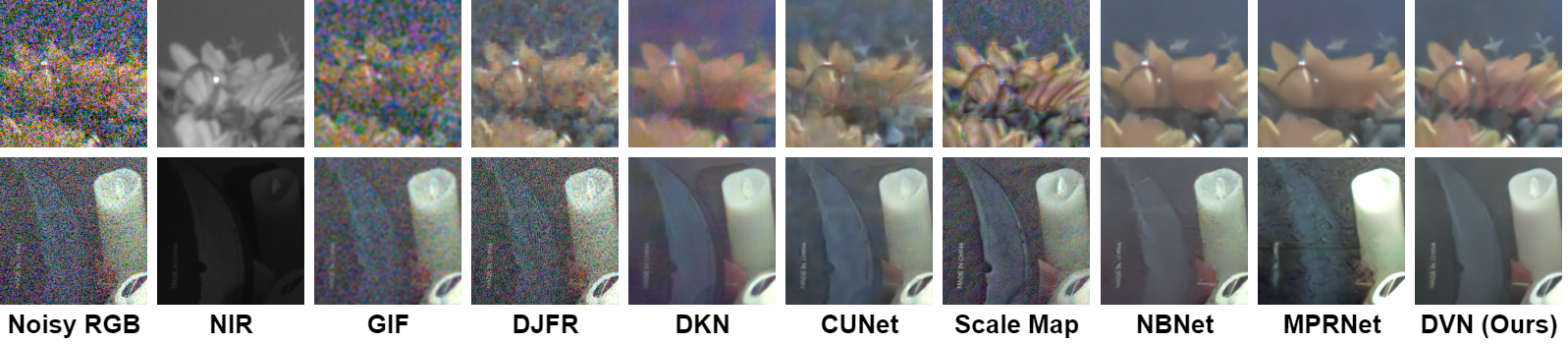}
	\caption{Fusion results on RGB-NIR image pairs with \textit{real} noise. DVN obviously obtains better results than other algorithms. }
	\label{fig:realCompare}
\end{figure*} 

\begin{figure*}[!t]
	\centering
	\includegraphics[width=0.78\linewidth]{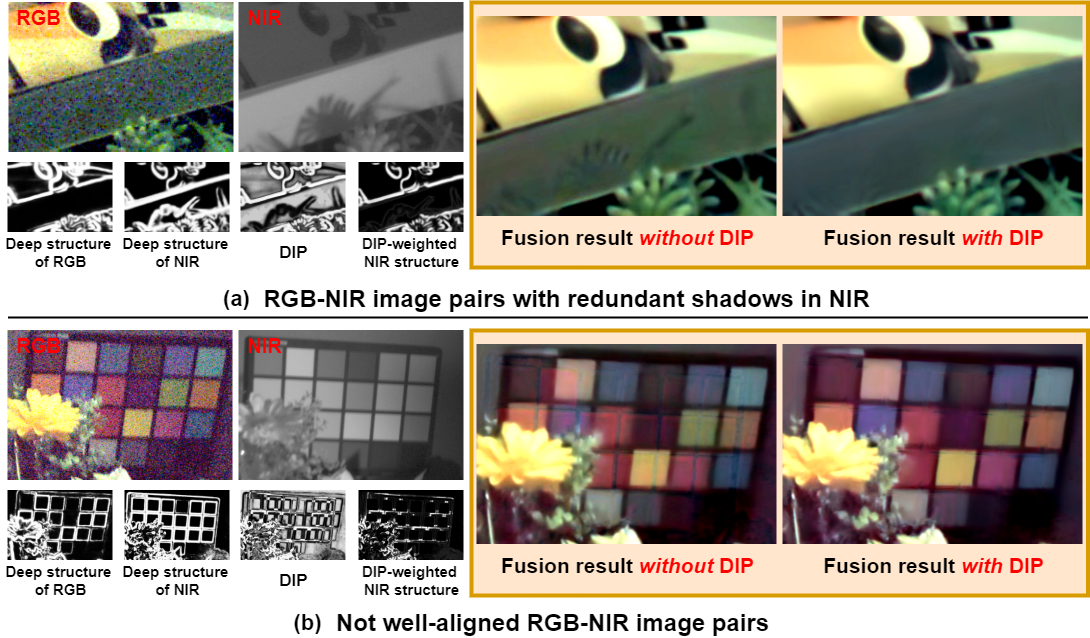}
	\caption{Illustration of the effectiveness of DIP. (a) shows a typical case of structure inconsistency caused by NIR shadows and (b) shows a case of the misalignment RGB-NIR. Fusion results and visualizations of deep structures verify the effectiveness of the DIP. Both examples are gathered from real noisy image pairs. }
	\label{fig:effectivedip}
\end{figure*}

\subsubsection{Generalization on Real Noisy RGB-NIR. }
To evaluate the performance of algorithms when facing \textit{real} low-light images, we conduct a qualitative experiment on several pairs of RGB-NIR images captured in real low-light environments.
As shown in Figure \ref{fig:realCompare}, outputs of DVN have obviously low noise, rich details, and are visually more natural when handling RGB-NIR pairs with \textit{real} noise, even if the network is trained on a synthetic noisy dataset. 

\subsubsection{Comparison on Public Dataset. }
So far, there is no \textbf{high-quality} public RGB-NIR dataset like DVD yet. For example, RGB-NIR pairs in IVRG \cite{brown2011multi} are not well aligned. Even so, we retrained DVN and other methods on IVRG and give quantitative comparison in Table \ref{table:lowlight}. It is clear that DVN still performs well. 

\subsubsection{Comparison with Low-Light Enhancement Methods. }
We also compare our method with the low-light enhancement methods. We retrained SID \cite{chen2018learning} and SGN \cite{gu2019self}, the comparison can be seen in Table \ref{table:lowlight}. It is clear that our proposed DVN still shows great superiority.

\subsection{Effectiveness of DIP}
In this section, we verify that the proposed DIP is effective in handling the mentioned structure inconsistency. For comparison, we retrain a baseline, which is the same as the proposed DVN only without the DIP module. As Figure \ref{fig:effectivedip}(a) shows, the NIR shadow of the grass still remains in the fusion result without DIP, but not in the fusion result with DIP. This directly proves that DIP can handle the structure inconsistency. Figure \ref{fig:effectivedip}(b) shows that DIP can also deal with serious structure inconsistency caused by the misalignment between RGB-NIR images to a certain extent (this example pair cannot be aligned even after image registration). This has practical value because the problem of misalignment frequently occurs in applications. Taking into account the nature of DIP, the remaining artifacts are in line with expectations, since they are concentrated near the pixels 
with gradients in the RGB image. 

In addition, Figure \ref{fig:effectivedip} also visualizes the deep structure of RGB, NIR, consistent NIR (DIP-weighted) as well as DIP Maps. It is obvious that even facing noisy input, the RGB deep structure still contains clear structures. The visual comparison between the NIR deep structure and the \textbf{consistent} NIR deep structure proves that the introduction of DIP can handle structure inconsistency in deep feature space.
\vspace{-0.5em}
\subsection{Ablation Study}
We evaluate the effectiveness of each component in the proposed algorithm on the DVD benchmark quantitatively in this section. PSNR and SSIM are reported in Table \ref{table:ablation}. The baseline network directly fuse NIR features with RGB features (row 1 in Table \ref{table:ablation}). 

\begin{table}[!t]
\centering
\begin{tabular}{lccc|cc}
\hline
row. & $\mathcal{L}_{rec}^{\hat{C}} + \mathcal{L}_{rec}^{N}$ & DSEM & DIP & PSNR  & SSIM   \\ \hline \hline
1. & --             & --   & --  & 28.87 & 0.9356 \\ \hline
2. & $\checkmark$   & --   & --  & 29.30 & 0.9375 \\ \hline \hline
3. & --             & $\checkmark$ & --  & 29.06 & 0.9376 \\ \hline
4. & $\checkmark$   & $\checkmark$ & --  & 29.36 & 0.9358 \\ \hline
5. & $\checkmark$   & $\checkmark$ & $\checkmark$ & \textbf{29.62} & \textbf{0.9400} \\ \hline
\end{tabular}
\vspace{-0.5em}
\caption{Ablation experiment results are conducted on DVD to study the effectiveness of each component. $\sigma$ is set to 4. }
\label{table:ablation}
\end{table}

Intermediate supervision $\mathcal{L}_{rec}^{\hat{C}}$ and $\mathcal{L}_{rec}^{N}$ effectively improve the performance as Table \ref{table:ablation} (row 1 and 2) shows.
This indicates the necessity of enhancing the noise suppression capability of the network for clean structure extraction. 

Applying DSEM to learn deep structures without DIP can improve performance as well as Table \ref{table:ablation} (row 1 and 3) shows. However, since the inconsistent structures are not removed, the benefits are not obvious, even if we use intermediate supervision and DSEM simultaneously as row 4 shows.

As Table \ref{table:ablation} (row 5) shows, after introducing DIP to deal with the structure inconsistency, the network performance can be further improved by a large margin. This demonstrates the effectiveness of our proposed algorithm and the necessity to focus on the structure inconsistency problem on RGB-NIR fusion problem.

\section{Conclusion}
In this paper, we propose a novel RGB-NIR fusion algorithm called Dark Vision Net (DVN). DVN introduces Deep inconsistency prior (DIP) to integrate the structure inconsistency into the deep convolution features, so that DVN can obtain a high-quality fusion result without visual artifacts. In addition, we also proposed the first available benchmark, which is called Dark Vision Dataset (DVD), for RGB-NIR fusion algorithms training and evaluation. Quantitative and qualitative results prove that the DVN is significantly better than other algorithms.

\section{Acknowledgements}
This paper is supported by the National Key R\&D Plan
of the Ministry of Science and Technology (Project No.
2020AAA0104400) and Beijing Academy of Artificial Intelligence (BAAI).

\bibliography{aaai22}

\end{document}